# Contactless 3D Human Body Measurement Using Depth Cameras for Smart Health Monitoring

1st Martha Asare
*Department of Computer Science*
*University of Texas Rio Grande Valley*
Edinburg, USA
martha.asare01@utrgv.edu

2nd Xuan Wang
*Department of Information Systems*
*University of Texas Rio Grande Valley*
Edinburg, USA
xuan.wang@utrgv.edu

3rd Juan Lopez Alvarenga
*School of Medicine, Primary & Community Care*
*University of Texas Rio Grande Valley*
Edinburg, USA
juan.lopezalvarenga@utrgv.edu

4th Lois Akosua Serwaa
*Department of Human Genetics*
*University of Texas Rio Grande Valley*
Edinburg, USA
loisakosua.serwaa01@utrgv.edu

5th Jinghao Yang
*Department of Electrical and Computer Engineering*
*University of Texas Rio Grande Valley*
Edinburg, USA
jinghao.yang@utrgv.edu

***Abstract*—Contactless body measurement technologies are becoming increasingly significant for smart health monitoring, digital health applications, and remote patient assessment. Traditional anthropometric measurements typically necessitate physical contact and trained personnel, which may constrain scalability in remote healthcare settings. In this study, we introduce a depth camera-based framework for estimating human body measurements utilizing 3D point cloud data. An Orbbec Astra 2 depth camera was employed to capture RGB images, depth maps, and 3D point clouds of participants. The captured point cloud was processed using Python-based tools, including Open3D, NumPy, and OpenCV, to segment the human body from the background. Key anthropometric measurements, such as height and arm span, were computed. The measurements were obtained through a combination of spatial filtering and landmark selection on the 3D point cloud, followed by the projection of the computed measurements onto the corresponding RGB image using camera intrinsic parameters. In addition to linear measurements, the approximate body volume and visible surface area were estimated using voxel-based occupancy analysis and mesh-based surface reconstruction methods. The experimental results from a single depth capture demonstrated that accurate body measurements and geometric estimates could be obtained from depth camera data without physical contact. This study provides a foundation for future real-time systems that integrate depth sensing with intelligent health monitoring and generative AI models for smart healthcare applications.**

***Keywords—anthropometric measurement, depth cameras, generative AI, point cloud, smart health***

## I. Introduction

Accurate human body measurements are important in a wide range of healthcare and biomedical applications, including clinical assessment, rehabilitation monitoring, ergonomic evaluation, and fitness analysis[1,2]. Anthropometric parameters, such as height, arm span, body volume, and body surface area, provide useful indicators for evaluating physical health, nutritional status, and physiological development[3,4].

In recent years, large-scale anthropometric datasets have become increasingly significant for data-driven healthcare systems and emerging artificial intelligence applications. These structured body measurements can facilitate digital human modeling, health analytics, and predictive healthcare systems. Traditional anthropometric measurement methods typically rely on manual tools such as measuring tapes or stadiometers and require direct physical interaction with the subject[5]. These approaches may be time-consuming, require trained personnel, and may not be suitable for remote or continuous health monitoring scenarios[6].

Recent advancements in depth-sensing technologies have facilitated novel opportunities for contactless human-body measurement[7,8]. Depth cameras possess the capability to capture three-dimensional (3D) geometric data of a scene, including the shape and structure of the human body[9]. In contrast to conventional RGB cameras, depth sensors offer additional spatial information that enables the reconstruction of point clouds representing the 3D geometry of observed objects[10]. These capabilities have been extensively explored in applications such as motion tracking, human pose estimation, and 3D scene reconstruction[11,12]. In smart-healthcare environments, depth-sensing technologies can facilitate non-intrusive monitoring systems that capture anthropometric data without necessitating wearable sensors or direct contact[13]. By processing depth data and point clouds, it becomes feasible to estimate body measurements automatically and visualize these measurements directly on captured images[14]. Such sensing systems can support emerging smart-health applications, including remote patient monitoring, telemedicine assessment, and digital health analytics[15,16].

In this study, we examined the application of an Orbbec Astra 2 depth camera to estimate human body measurements

This work was supported by the National Science Foundation (NSF) CREST Center for Multidisciplinary Research Excellence in Cyber-Physical Infrastructure Systems (MECIS) under Award No. 2112650, NSF ExpandAI PARTNER: AI Research and Innovation for Smart Environments (ARISE) under Award No. 2434916, and NSF ACCESS: AI-Enhanced Cross-Scale Sensing and Intelligent Quality Assessment for Scalable Additive Manufacturing under Award No. ELE250047.

from 3D point cloud data. A Python-based processing pipeline was developed to capture RGB images, depth maps, and aligned point clouds, isolate the human subject from the surrounding environment, and compute key anthropometric measurements, including height and arm span. Additionally, the approximate body volume and visible surface area were estimated using voxel-based occupancy analysis and mesh-based surface reconstruction techniques[17,18]. The resulting measurements were projected onto the corresponding RGB image using camera intrinsic parameters for visualization and verification.

This study makes several significant contributions. It develops a contactless body measurement pipeline utilizing an Orbbec depth camera and point cloud processing. It estimates key anthropometric measurements, including height and arm span, from 3D point cloud data. Additionally, it provides an approximate estimation of body volume and visible surface area through voxel-based and mesh-based reconstruction methods. Furthermore, it demonstrates the feasibility of depth-camera-based sensing for smart health-monitoring applications.

## II. Methodology

### A. Conceptual Framework

The proposed framework facilitates the estimation of human body measurements from depth-camera data through a multistage processing pipeline. As shown in Fig. 1, the system comprises four primary stages: data acquisition, point cloud processing and person segmentation, anthropometric measurement estimation, and geometric analysis for volume and surface area estimation. Initially, RGB images, depth maps, and 3D point clouds were captured using an RGB-D camera.

The acquired point cloud was subsequently processed to isolate the human subject from the background using spatial filtering and clustering techniques. Following segmentation, anthropometric measurements, such as height and arm span, were calculated from the selected anatomical landmarks. Finally, voxel-based occupancy analysis and mesh reconstruction methods were employed to estimate the approximate body volume and visible surface area. This framework enables the contactless extraction of body measurements from depth camera data and supports potential smart health monitoring applications based on three-dimensional sensing technologies [13, 19].

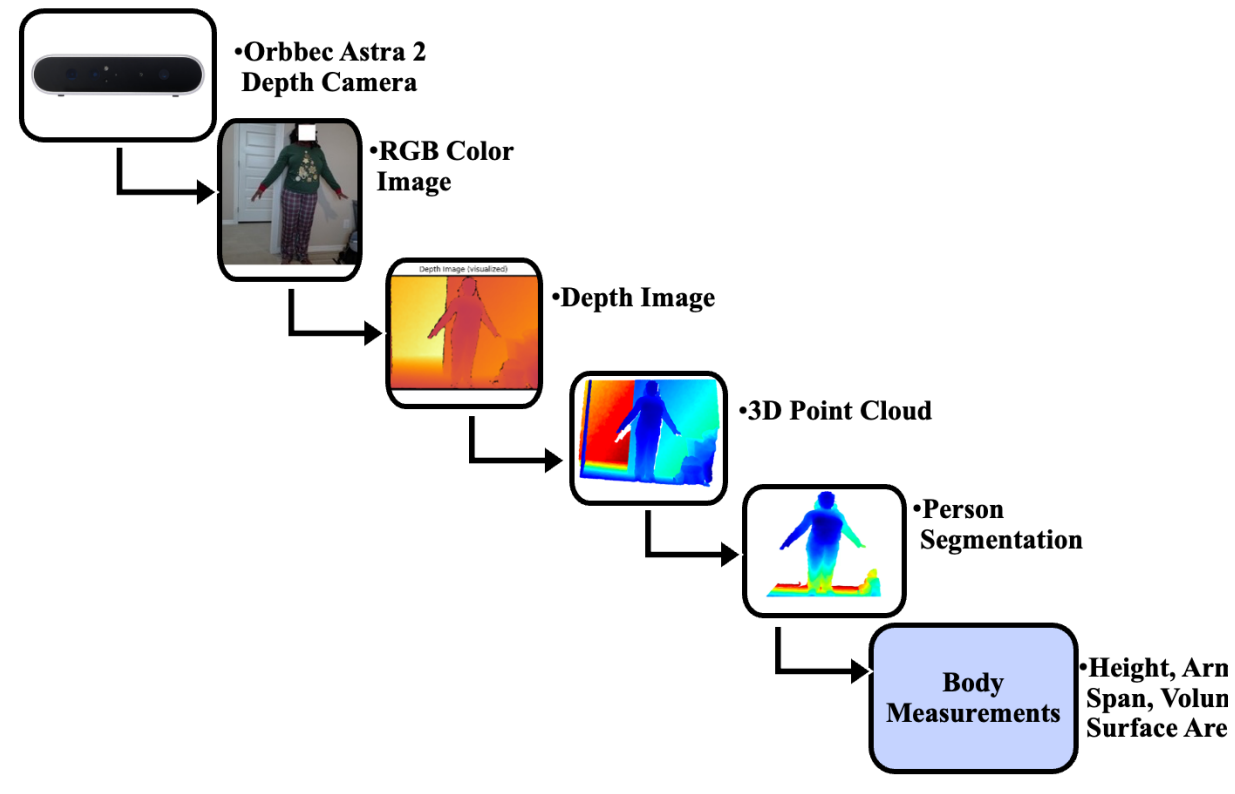

Fig. 1. Proposed pipeline for contactless human body measurement using depth-camera data. RGB and depth images were captured using an Orbbec Astra 2 sensor. The depth image was converted into a 3D point cloud, followed by person segmentation to isolate the human subject. Anthropometric measurements, including height, arm span, body volume, and visible surface area, were estimated from the segmented point cloud.

### B. Data Acquisition

Depth data were acquired utilizing an Orbbec Astra 2 RGB depth camera, which employs a structured-light depth sensor capable of capturing synchronized RGB images, depth maps, and three-dimensional (3D) point cloud data[20]. RGB depth sensors are extensively utilized for 3D reconstruction and human sensing because of their ability to provide geometric information about objects in the environment in real time [21,22]. The Astra 2 sensor functions within a depth range of approximately 0.6–8 m and supports high-resolution RGB and depth capture [20].

The RGB stream was captured at a resolution of 1920 × 1080 pixels at 30 frames per second, while the depth stream generated a 16-bit depth map representing the distance between each pixel and the camera sensor. Utilizing the Orbbec Viewer software, the captured data were exported as RGB images, depth images, and a 3D point cloud in PLY format. Each point in the point cloud contains real-world coordinates $(X, Y, Z)$ expressed in millimeters relative to the camera coordinate system, where the X-axis denotes the horizontal direction, the Y-axis denotes the vertical direction, and the Z-axis denotes the depth from the camera optical center.

To enable accurate projection between the 3D point cloud and the RGB image, the intrinsic calibration parameters of the camera were obtained from the Orbbec configuration file. The projection of a 3D point in camera coordinates onto the 2D image plane conforms to the pinhole camera model [23,24], as shown in (1).

$$s\begin{bmatrix} u \\ v \\ 1 \end{bmatrix} = K[R \quad t]\begin{bmatrix} X \\ Y \\ Z \\ 1 \end{bmatrix}, \tag{1}$$

where $(u, v)$are 2D image coordinates; $(X, Y, Z)$ are 3D world coordinates; $K$ is intrinsic matrix; $R, t$ are rotation and translation matrices (extrinsic) and $s$ is a scalar factor. The intrinsic parameters include the focal length $(f_x, f_y)$, and the optical center$(c_x, c_y)$, also known as the principal point. The intrinsic camera matrix K is defined as:

$$K = \begin{bmatrix} f_x & 0 & c_x \\ 0 & f_y & c_y \\ 0 & 0 & 1 \end{bmatrix}$$

For the Orbbec Astra 2 camera used in this study, the intrinsic matrix K is defined as follows:

$$K = \begin{bmatrix} 1245.897 & 0 & 962.633 \\ 0 & 1246.242 & 533.425 \\ 0 & 0 & 1 \end{bmatrix}$$

These intrinsic parameters enable the projection of 3D points onto an RGB image for visualization and verification of the computed anthropometric measurements.

### C. Point Cloud Processing and Person Segmentation

The exported point cloud was processed using a Python-based pipeline utilizing the Open3D library, which offers efficient tools for point cloud manipulation and visualization [13]. Initially, the .ply point cloud file generated by the Orbbec Viewer was loaded and converted into a NumPy array containing a set of 3D coordinates representing the captured scene. The raw point cloud included the human participant and surrounding background objects, such as walls and furniture. To isolate the human body, a segmentation procedure was employed.

The point cloud was first downsampled to reduce the computational complexity while preserving the geometric structure of the scene. Subsequently, manual landmark selection was conducted using Open3D's interactive point-picking interface. Four anatomical landmarks were selected: the top of the head, bottom of the feet, left hand, and right hand. These landmarks delineate a spatial region of interest (ROI) enclosing the participant. Points outside the defined region were removed to eliminate background elements[13, 25].

A clustering-based filtering step was then applied to retain the largest connected component corresponding to the human body. The resulting segmented point cloud represents the person-only 3D geometry and is utilized for subsequent anthropometric measurement estimation. The segmentation process from the original point cloud to the isolated human body is shown in Fig. 2.

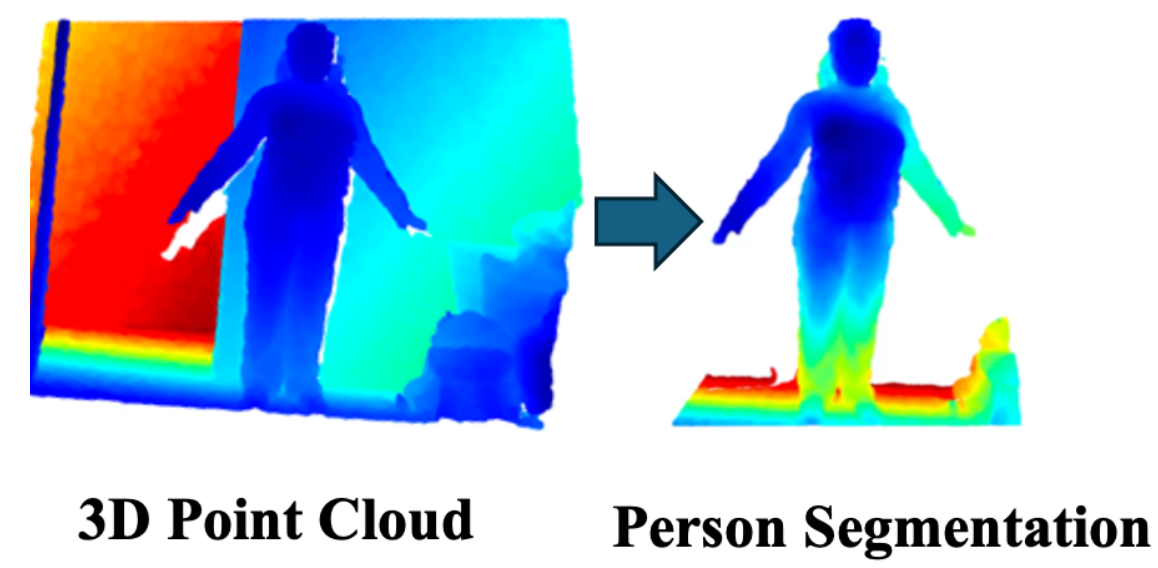

Fig. 2. Point cloud processing and person segmentation. The raw 3D point cloud captured from the depth camera contains both human subjects and the surrounding environment. Segmentation techniques are applied to isolate the human body from background objects for further anthropometric analysis.

## III. Experimental Results

This section outlines the experimental findings obtained using the proposed contactless body measurement pipeline, which leverages depth-camera sensing technology. Utilizing an Orbbec Astra 2 depth camera, RGB images, depth maps, and 3D point cloud data were captured and subsequently processed according to the methodology detailed in Section 2. The segmented point cloud of the human participant was instrumental in calculating essential anthropometric measurements and geometric properties of the body.

The experiments were primarily focused on two key outputs of the system. The first output involved estimating anthropometric measurements, such as human height and arm span, by identifying specific anatomical landmarks within the segmented point cloud. The second output entailed conducting a geometric analysis to approximate body volume and visible body surface area by employing voxel-based occupancy and mesh reconstruction techniques.

The following subsections provide a comprehensive analysis of the results pertaining to the measurement estimation and geometric reconstruction derived from the captured dataset.

### A. Anthropometric Measurement Estimation

Anthropometric measurements encompass quantitative assessments of the human body, detailing physical dimensions such as height, limb length, and body proportions [26, 27]. These measurements are extensively utilized in healthcare, clinical evaluation, ergonomics, and studies of human physiology to assess body structure and physical development[27]. In the context of smart health monitoring systems, automated anthropometric measurement techniques facilitate non-contact estimation of body dimensions using sensing technologies and computational analysis.

In this study, anthropometric measurements were derived from a segmented 3D point cloud generated by a depth-camera capture. Following the isolation of the human participant from the surrounding environment, key anatomical landmarks were identified through an interactive point selection process. Four landmarks, corresponding to the top of the head, bottom of the feet, left hand, and right hand, were selected using the Open3D point-picking interface. These landmarks delineate the spatial boundaries of the human body and facilitate the estimation of body dimensions directly from the 3D geometry.

Anthropometric measurements were computed using the Euclidean distance between the selected 3D landmark points[13, 28, 29]. The distance between two points in 3D space is defined as follows:

$$d = \sqrt{(x_2 - x_1)^2 + (y_2 - y_1)^2 + (z_2 - z_1)^2} \quad (2)$$

where $(x_1, y_1, z_1)$ and $(x_2, y_2, z_2)$ represent the coordinates of two points in the point cloud.

Using the formula in (2), human height was estimated as the distance between the head and feet landmarks, whereas arm span was computed as the distance between the left and right hand landmarks. These measurements are derived directly from the reconstructed 3D geometry of the participant and therefore represent real-world body dimensions.

For the captured dataset, the estimated anthropometric measurements were as follows:

- Height: 151.6 cm
- Arm Span: 61.4 cm

To confirm the accuracy of the estimated measurements, the chosen 3D landmark points were mapped onto the corresponding RGB image by applying the camera intrinsic parameters outlined in Section 2. The resulting measurements were then overlaid onto the RGB image to facilitate visualization and verification, as shown in Fig. 3.

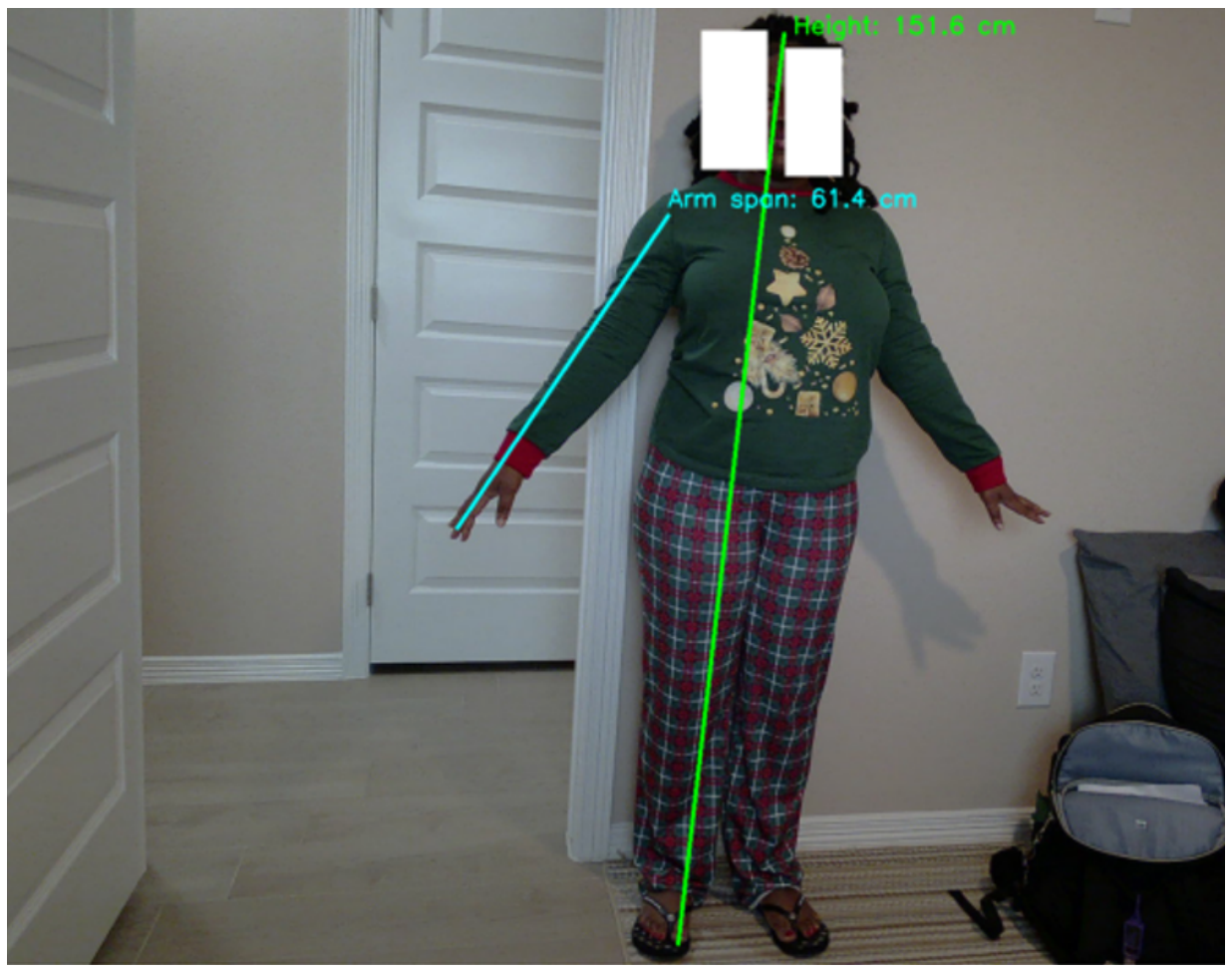

Fig. 3. Estimation of anthropometric measurements from a segmented point cloud. Key anatomical landmarks, including the top of the head, bottom of the feet, and both hands, were identified to calculate body measurements. The estimated height and arm span were projected onto an RGB image using the camera intrinsic parameters for visualization and verification.

### *B. Body Volume and Surface Area Estimation*

In addition to linear anthropometric measurements, geometric properties, such as body volume and body surface area, are crucial indicators for evaluating health and analyzing body composition. These metrics can be derived from three-dimensional point cloud data captured by depth sensors [30, 31]. After segmentation, the isolated human body point cloud consisted of approximately 338,254 points, depicting the subject's visible geometry.

To calculate body volume, the segmented point cloud was transformed into a voxel grid, where the space is partitioned into small cubic units termed voxels. Voxel-based models are widely utilized for estimating the volume of irregular 3D forms due to their straightforward occupancy-based approximation [18]. A voxel resolution of 8 mm × 8 mm × 8 mm was chosen to balance geometric precision with computational efficiency in human-scale reconstruction. Consequently, each voxel represents a cubic volume of $V_{voxel} = (8mm)^3 = 512mm^3$

A voxel is deemed occupied if at least one point from the segmented point cloud is located within the voxel. Let $N_{voxel}$ denote the number of occupied voxels. The total volume is estimated as:

$$V = N_{voxel} \times V_{voxel} \quad (3)$$

For the dataset in this study, the segmentation process resulted in approximately 27,333 occupied voxels. By substituting these values, we obtain

$$V = 27333 \times 512 \approx 1.399 \times 10^{10} mm^3$$

Converting cubic millimeters to liters ($1L = 10^6\ mm^3$) Gives $V \approx 13.99L$ Thus, the estimated occupied body volume from the captured dataset is approximately 13.99 liters.

To estimate the body surface area, a triangular mesh was reconstructed from the segmented point cloud using an alpha-shape surface reconstruction method, which approximates the outer boundary of the point set [32]. The reconstructed mesh employed for the geometric analysis is shown in Fig. 4.

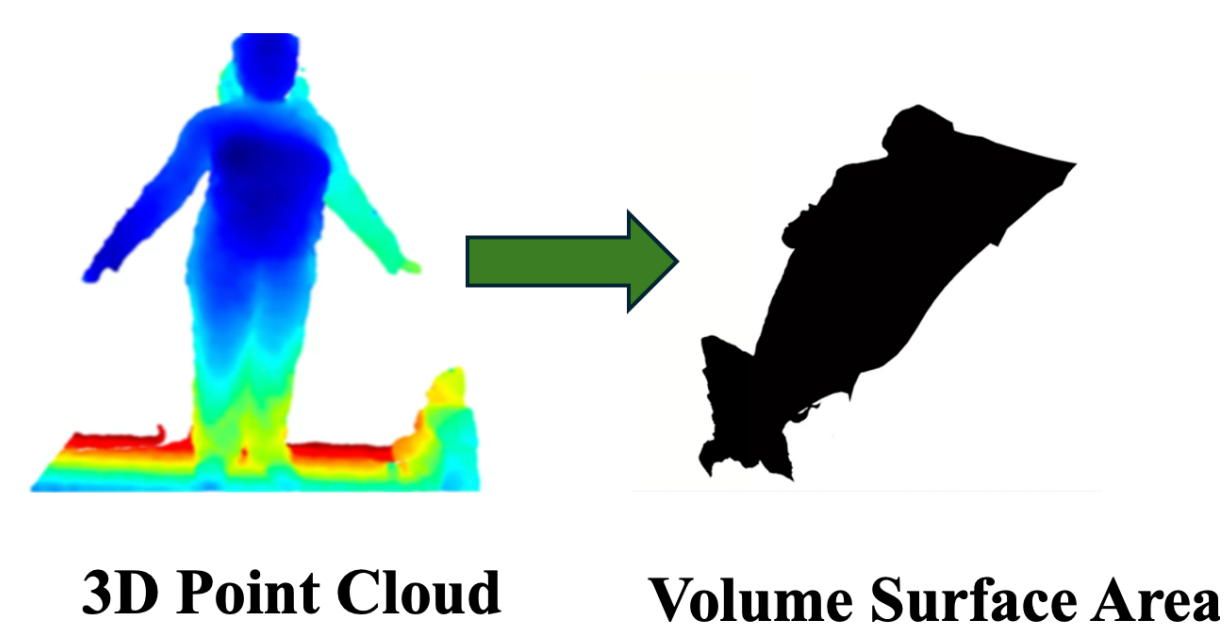

Fig. 4. Reconstructed triangular mesh of the segmented human body point cloud used for the estimation of body volume and surface area.

The surface area was computed by summing the areas of all triangular faces in the reconstructed mesh:

$$A = \sum_{i=1}^{n} A_i \quad (4)$$

where $A_i$ represents the area of triangle $i$. The area of each triangle is calculated from its three vertices using the cross-product formulation

$$A = \frac{1}{2} \|(v_2 - v_1) \times (v_3 - v_1)\| \quad (5)$$

where $v_1, v_2$ and $v_3$ denote the 3D coordinates of the triangle vertices. By summing the areas of all triangles in the reconstructed mesh, the estimated visible body surface area was obtained as $A \approx 3.03m^2$.

Given that the depth data were acquired from a single viewpoint, the reconstructed mesh represents only the visible segment of the body surface. Consequently, the estimated surface area pertains to the visible geometry rather than the entire anatomical surface.

## IV. Conclusion and Future Works

This study introduces a framework for contactless human body measurement utilizing depth-camera sensing and 3D point cloud processing, aimed at applications in smart health monitoring. The proposed framework generates structured 3D anthropometric data that can serve as a foundational dataset for generative AI models in smart health applications. These models have the potential to facilitate the creation of personalized body models, digital human representations, and predictive health analytics, thereby supporting the development of intelligent and scalable smart health monitoring systems.

Employing an Orbbec Astra 2 depth camera, RGB images, depth maps, and aligned point cloud data were captured and processed to estimate key anthropometric measurements. A Python-based processing pipeline was developed to isolate the human body from the surrounding environment, extract anatomical landmarks, and compute body dimensions directly from the reconstructed three-dimensional geometry.

The experimental results indicate that significant anthropometric parameters, such as human height and arm span, can be estimated from depth-derived point cloud data using geometric distance computation. In addition to linear measurements, the framework facilitates the approximate estimation of body volume and visible body surface area through voxel-based occupancy analysis and mesh-based surface reconstruction. These findings demonstrate the potential of depth-camera sensing as a viable solution for contactless body measurement in smart health-monitoring environments, where traditional manual measurement methods may be inconvenient or impractical.

Despite these promising results, this study has some limitations. The current implementation relies on single-view depth capture, which restricts the completeness of the reconstructed body geometry and results in partial estimates of body volume and surface area. Additionally, factors such as occlusions, clothing variations, and depth sensor noise may affect the segmentation quality and geometric reconstruction accuracy.

Future work will focus on extending the proposed framework to real-time body measurement using the PyOrbbec SDK on a Linux platform, thus enabling automated and continuous anthropometric monitoring. In addition, multi-view depth acquisition and improved segmentation techniques will be investigated to enhance the completeness and accuracy of the reconstructed body geometry.

## Acknowledgment

The authors thank the University of Texas Rio Grande Valley for providing the research environment and computational resources that supported this study. The authors appreciate the valuable guidance and discussions that contributed to the development of this research.